\documentclass[lettersize,journal]{IEEEtran}
\usepackage{amsmath,amsfonts}
\usepackage{algorithmic}
\usepackage{algorithm}
\usepackage{array}
\usepackage{caption}
\usepackage[caption=false,font=normalsize,labelfont=sf,textfont=sf]{subfig}
\usepackage{textcomp}
\usepackage{stfloats}
\usepackage{url}
\usepackage{verbatim}
\usepackage{booktabs}
\usepackage{multirow}
\usepackage{makecell}
\usepackage[table]{xcolor}
\usepackage{graphicx}
\usepackage{subcaption}
\usepackage{cite}
\begin{document}
\title{TAR: Text Semantic Assisted Cross-modal Image Registration Framework for Optical and SAR Images}
\author{Zhuoyu Cai, Dou Quan,~\IEEEmembership{Member,~IEEE}, Ning Huyan,~\IEEEmembership{Member,~IEEE}, Pei He,~\IEEEmembership{Member,~IEEE}, Shuang Wang,~\IEEEmembership{Senior Member,~IEEE}, Licheng Jiao,~\IEEEmembership{Life Fellow, IEEE}
		\thanks{Manuscript created May 2026; This work was supported in part by the National Natural Science Foundation of China under Grant 62201407, 62501356, and 62271377; in part by the Key Research and Development Program of Shanxi Program under Grant 2023QCYLL28, Grant 2024GX-ZDCYL-02-08, and Grant 2024GX-ZDCYL-02-17; in part by the China Postdoctoral Science Foundation under Grant 2022M722496; and in part by the Key Scientific Technological Innovation Research Project by Ministry of Education. (Corresponding author: Dou Quan, Ning Huyan, e-mail: dquan@stu.xidian.edu.cn; quandou@xidian.edu.cn; n-hy@mail.tsinghua.edu.cn.)}
		\thanks{Zhuoyu Cai, Dou Quan, Shuang Wang, Pei He, and Licheng Jiao are with the Key Laboratory of Intelligent Perception and Image Understanding of Ministry of Education of China, School of Artificial Intelligence, Xidian University, Xi'an 710071, China. Ning Huyan is with the Department of Automation, Tsinghua University, Beijing 100084, China.}}
        
\markboth{IEEE Transactions on Geoscience and Remote Sensing, May~2026}%
{TAR: Text Semantic Assisted Cross-modal Image Registration Framework for Optical and SAR Images}


\maketitle

\begin{abstract}
Existing deep learning-based methods can capture shared features from optical and synthetic aperture radar (SAR) images and align them in space. However, their registration accuracy still needs to be further improved when there are large geometric deformations between cross-modal images, as the deep model has difficulty simultaneously handling cross-modal appearance discrepancies and complex spatial transformations. To address this issue, this paper proposes a text semantic-assisted cross-modal image registration framework, named TAR, for optical and SAR images. TAR aims to exploit text semantic priors in remote sensing images to alleviate the modality gap between optical and SAR images and enhance cross-modal image feature learning, thereby improving the accuracy of cross-modal image matching. TAR mainly contains a multi-scale visual feature learning (MSFL) module, a text-assisted feature enhancement (TAFE) module, and a coarse-to-fine dense matching (CFDM) module. Firstly, the MSFL module extracts multi-scale visual features from optical and SAR images, respectively. Secondly, the TAFE module leverages text semantic features to enhance the feature consistency of cross-modal images and improve the feature discriminability for more accurate matching through visual-text feature interaction. Specifically, it builds text descriptors that are related to remote sensing image scenes and land-cover objects, and then uses frozen RemoteCLIP to extract text features from these text descriptors. After that, TAFE enhances the image features through visual-text feature interaction. Finally, the CFDM module performs coarse matching based on the enhanced high-level image features and conducts fine matching based on low-level image features. Experimental results on cross-modal remote sensing images demonstrate the effectiveness and advantages of the proposed TAR framework, which outperforms several state-of-the-art approaches and yields more significant performance gains under large geometric deformations.
\end{abstract}


\section{Introduction}
\IEEEPARstart{M}{ultimodal} remote sensing data have been widely used in intelligent Earth observation \cite{zhu2017deep, ghassemian2016review}, such as optical and synthetic aperture radar (SAR) images, which provide complementary information in a complex imaging environment \cite{schmitt2018sen1}. Specifically, optical images contain rich spectral and texture details, while SAR images can be acquired under all weather conditions and reflect the scattering properties of ground objects. Optical and SAR image registration is essential for their fusion applications \cite{zitova2003image, zhu2024multimodal}, such as object detection \cite{li2020object}, change detection \cite{shafique2022deep}, and land-cover mapping classification. However, optical and SAR image registration remains challenging because there are significant differences in image intensity, texture, and local structures between multimodal images caused by various imaging mechanisms \cite{ye2017robust}. The core of cross-modal remote sensing image registration is to extract shared features under strong appearance differences.


\begin{figure}[t]
    \centering
    \includegraphics[width=1.0\linewidth]{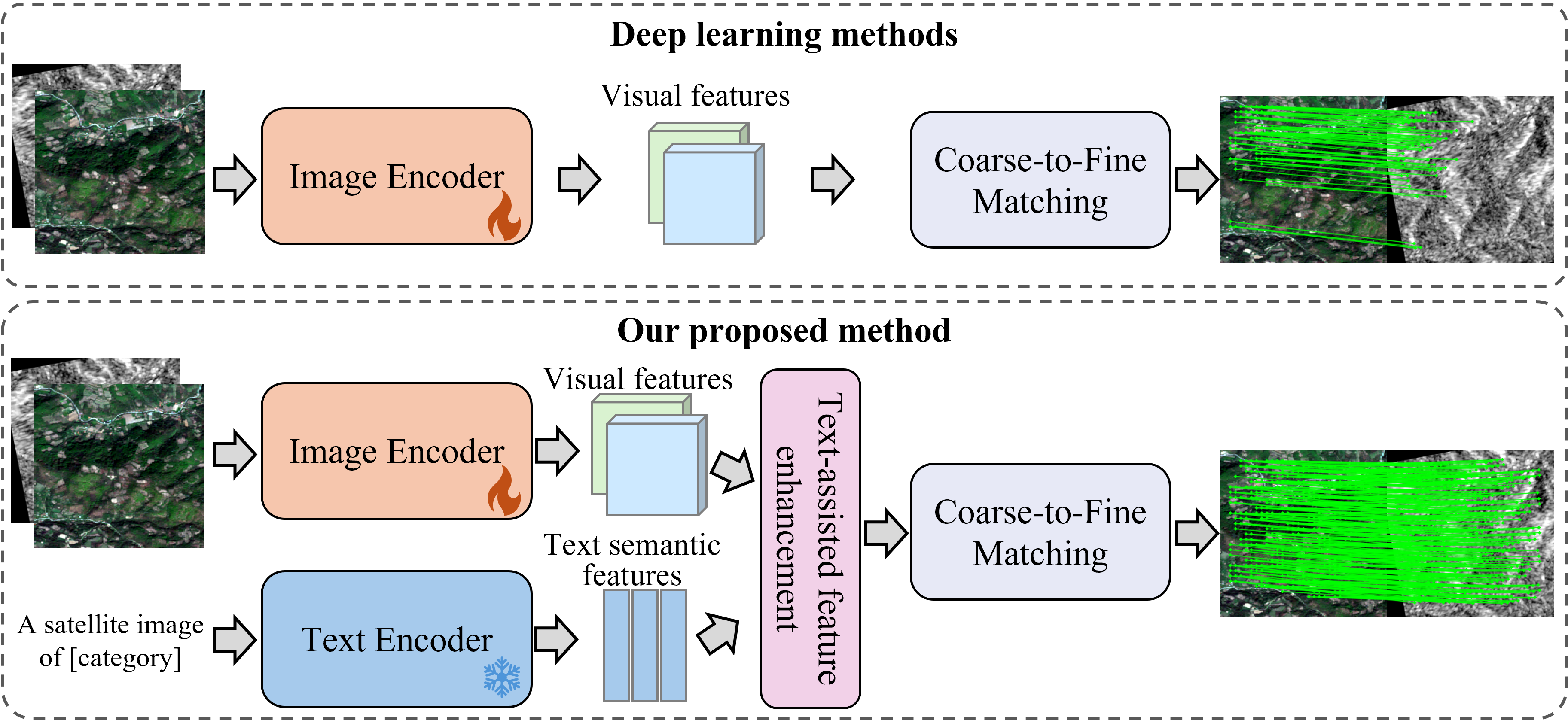}
    \caption{Deep learning methods for image registration and our proposed text semantic-assisted image registration framework.} 
    \label{figintro}
\end{figure}


Recently, deep learning-based methods have shown significant advantages over traditional handcrafted feature-based methods on image registration \cite{li2025deep}. Deep learning-based methods usually adopt a deep convolutional network \cite{quan2022deep,quan2023efficient,quan2022self} or a Transformer \cite{sun2021loftr, wang2022matchformer, chen2022aspanformer, edstedt2023dkm, edstedt2024roma} to learn high-level visual representations from images, which are robust to illumination changes, geometric transformation, and modality changes. After that, deep learning-based methods build matching correspondences through feature similarity and then perform image transformation for image alignment in space. However, existing deep learning-based methods mainly rely on visual features for image matching and registration, as shown in Fig. \ref{figintro}. When the remote sensing images have obvious differences in appearance and large geometric deformations, their registration accuracy can decrease significantly.  The main reason is that the deep model needs to simultaneously deal with the cross-modal appearance discrepancies and complex spatial transformations. For cross-modal images with large geometric deformations, the deep model needs to search for matching correspondences from a wide spatial range while maintaining feature invariance across modalities and discrimination across local patterns, which increases ambiguity in the image matching. Thus, the visual features are insufficient for precise cross-modal image registration. 


To solve this problem, this paper proposes a Text semantic-Assisted image Registration framework, termed TAR, for optical and SAR images. TAR mainly introduces remote sensing semantic priors to enhance feature learning of cross-modal images and boost registration performance, as shown at the bottom of Fig. \ref{figintro}. It contains three main components: a Multi-Scale Visual Feature Learning (MSFL) module, a Text-Assisted Feature Enhancement (TAFE) module, and a Coarse-to-Fine Dense Matching (CFDM) module. Firstly, MSFL extracts multi-scale visual features from the optical and SAR images, respectively. Secondly, TAFE constructs text semantic prompts from common remote sensing scenes and land-cover categories and encodes them using a frozen RemoteCLIP text encoder \cite{liu2024remoteclip}. After that, TAFE performs visual feature enhancement through visual-text feature interaction and visual-visual interaction. The former leverages the text semantic information to enhance visual features, while the latter enhances the visual features through structural relationships within image modality and correlation across image modalities. Finally, we use the enhanced visual features for matching. Specifically, CFDM establishes initial correspondences using the enhanced high-level features and refines the matched locations using enhanced low-level local features. Experiments on optical and SAR images show that TAR achieves stronger matching performance than other representative registration methods, especially under precise localization requirements and large geometric deformations.

The main contributions of this paper are summarized as follows:
\begin{enumerate}
    \item We propose TAR, a text-assisted framework for optical and SAR image registration. Different from existing methods that mainly rely on visual features for matching, TAR introduces remote sensing semantic priors to enhance visual features for coarse matching, which can establish more reliable correspondences under large geometric deformations and significant appearance differences between cross-modal images.\\

    \item We design a Text-Assisted Feature Enhancement module, termed TAFE. TAFE builds a remote sensing text feature library from category prompts encoded by a frozen RemoteCLIP text encoder. It enhances high-level visual features through visual-text cross-attention and fuses the text-enhanced branch with the original visual interaction branch by an MLP, providing more informative coarse-level features for matching.\\

    \item We conduct extensive experiments on the SEN1-2 and OSdataset datasets to show the effectiveness and advantages of the text-assisted feature enhancement. The results show that TAR achieves the best correct match rates under multiple error thresholds while maintaining competitive registration errors. TAR particularly presents significant improvements under large geometric deformations.
\end{enumerate}

The remainder of this article is organized as follows. Section \ref{relatedwork} reviews related work on image registration. Section \ref{method} presents the proposed method. Section \ref{experiment} reports the experimental results and analysis. Section \ref{conclusion} presents the conclusion of this paper.






\section{Related Work}\label{relatedwork}
\subsection{Image Registration}
Image registration aims to establish spatial correspondences between two images acquired from different viewpoints,  at different times, or captured by various sensors \cite{zhang2025multi}.

Early registration methods mainly relied on handcrafted local features \cite{brown2005multi, hauagge2012image}, such as SIFT \cite{lowe2004distinctive} and SURF \cite{bay2006surf}. These traditional descriptors mainly rely on consistent intensity, texture, and gradient between images for matching, which fail to deal with significant image appearance differences between optical and SAR images. To improve robustness to image modality differences, HOPC \cite{ye2017robust} uses the structural information to build feature descriptors based on the histogram of oriented phase congruency. CFOG \cite{ye2019fast} adopts the oriented gradients of images for remote sensing image matching and registration. RIFT \cite{li2019rift} proposes a radiation-variation insensitive feature transform, which exploits phase congruency for feature point detection and uses a maximum index map for feature descriptions that are less sensitive to radiometric changes. LNIFT \cite{li2022lnift} proposes a locally normalized image feature transform, which reduces nonlinear radiometric differences between multimodal images through local normalization. However, their performance is limited in images with weak texture, complex scenes, and large geometric deformations \cite{ma2015robust}.
 
Deep learning methods utilize the learned feature representations to replace handcrafted features for matching and achieve significant advantages on image registration \cite{tyszkiewicz2020disk, luo2020aslfeat}. CNet \cite{quan2022deep} proposes a deep feature correlation learning for cross-modal remote sensing image registration. DW-Net \cite{quan2023efficient} introduces the wavelet transformation into deep features learning and improves the efficiency and robustness of cross-modal image registration. SD-Net \cite{quan2022self} proposes a self-distillation feature learning network for optical and SAR image registration, which leverages the rich feature similarity between samples for enhancing deep model optimization. GeoMamba \cite{cao2025efficient} uses a hierarchical Mamba network to extract multi-scale features for improving the efficiency and accuracy of remote sensing image registration. F3Net \cite{F3Net} adopts the adaptive frequency feature filtering network to explore useful frequency information for cross-modal remote sensing image registration. RSENet \cite{nie2024novel} generates  pseudo-optical images by an image translation network for optical and SAR image matching.  

Beyond local feature learning, some works propose end-to-end image registration methods. MU-Net \cite{ye2022multiscale} proposes a multi-scale framework for remote sensing image registration, which directly learns the mapping from the image pairs to the transformation parameters by an end-to-end method. 
ADRNet \cite{xiao2024adrnet} combines the rigid affine and non-rigid registration methods for aligning multimodal remote sensing images. GDROS \cite{sun2025gdros} proposes a geometry-guided dense registration framework for optical and SAR images, which uses the CNN-Transformer to extract multi-scale features. LoFTR \cite{sun2021loftr} directly predicts dense matching through a coarse-to-fine matching strategy, which captures feature relationships in single-modal images and cross-modal images by self-attention and cross-attention learning. RoMa \cite{edstedt2024roma} combines the pretrained model DINOv2 with fine-grained convolutional features to improve the robustness of dense feature matching. XoFTR \cite{tuzcuouglu2024xoftr} combines pre-training based on the masked image modeling and fine-tuning with pseudo image augmentation to deal with the image modality difference between thermal infrared and visible images.


However, existing deep learning methods only leverage image visual features for image registration. They are difficult to capture consistent and shared semantic features from cross-modal images when there are significant appearance differences and large geometric transformations between images. Although optical and SAR images have obvious appearance differences and geometric deformations, they have consistent semantic information about remote sensing scene categories and land-cover objects. Motivated by this, we introduce text semantic information to assist image visual feature learning, enhancing the robustness of visual features to image modality changes and geometric transformations, and achieving reliable matching correspondences for accurate optical and SAR image registration. 

\subsection{Textual Semantic Information in Vision Tasks}
Textual information provides semantic priors in vision tasks. Unlike image visual features, text descriptors contain more abstract information about objects, scenes, and attributes. CLIP \cite{radford2021learning} and ALIGN \cite{jia2021scaling} learn a shared vision-language representation space through large-scale image-text contrastive learning, which aligns the visual and natural language descriptions in feature space. Vision-language learning has been extended to the remote sensing images \cite{li2023rs, zhang2024rs5m}. RemoteCLIP \cite{liu2024remoteclip} adapts vision-language pretraining for various remote sensing tasks, such as zero-shot classification and cross-modal retrieval.

Textual semantics have also been introduced into various vision tasks, which can provide high-level semantic priors for enhancing the discrimination and robustness of visual features. For example, SegCLIP~\cite{zhang2024segclip} leverages the CLIP text encoder to extract text information, which guides visual features for distinguishing different classes in remote sensing semantic segmentation. TopicGEO~\cite{wang2025topicgeo} introduces the textual object semantics learned from CLIP into query-to-reference image matching and enhances spatial geolocation performance. TARCNet~\cite{he2025text} uses linguistic priors to enhance cross-modal consistency between hyperspectral image and light detection and ranging (LiDAR) data by contrastive learning. These studies demonstrate that text semantic priors can effectively guide and enhance visual feature learning in various remote sensing tasks.

Although text-assisted visual learning has been explored in classification, retrieval, and semantic understanding, there is little work to explore the effectiveness of the text semantic information in remote sensing image geometric correspondence estimation. In this paper, we try to leverage text semantic priors to enhance image visual feature learning, reducing the influence of cross-modal image differences and large geometric transformations. Text descriptions of remote sensing scene categories provide modality-agnostic semantic references, which can boost shared visual feature learning from cross-modal images and build reliable matching correspondences between optical and SAR images.

\section{Method} \label{method}

\begin{figure*}
    \centering
    \includegraphics[width=1\linewidth]{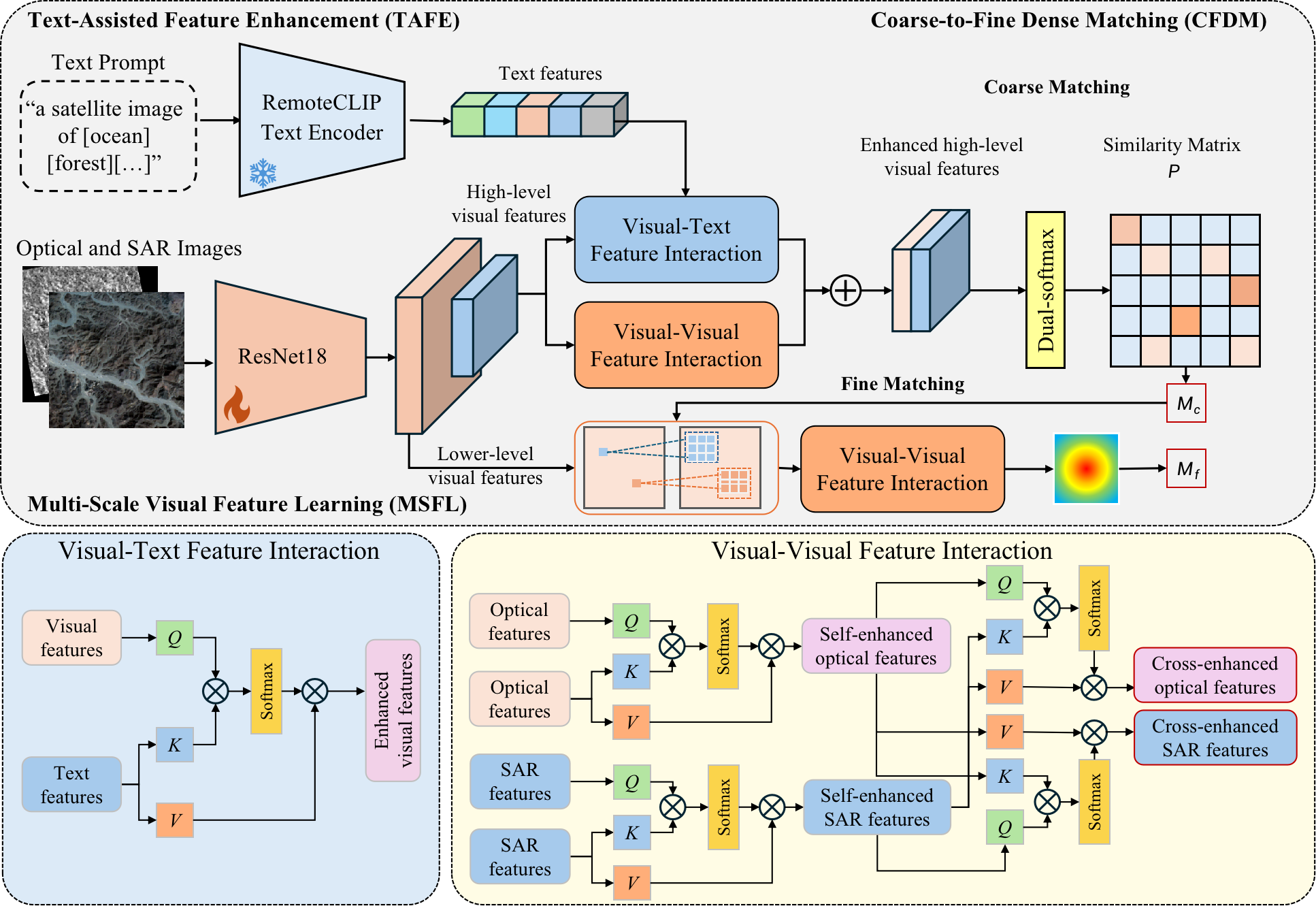}
    \caption{The overall framework of our proposed text semantic-assisted registration (TAR). TAR mainly contains three components: a Multi-Scale Visual Feature Learning (MSFL) module, a Text-Assisted Feature Enhancement (TAFE) module, and a Coarse-to-Fine Dense Matching (CFDM) module.}
    \label{fig:tar_framework}
\end{figure*}

The overall framework of our proposed text semantic-assisted registration (TAR) is shown in Fig. \ref{fig:tar_framework}. Unlike existing deep learning methods that rely solely on visual features for matching, TAR introduces remote sensing semantic text priors to enhance matching performance. TAR utilizes the remote sensing scene and land-cover information to reduce ambiguous matches, particularly in cross-modal images that have significant appearance differences and large spatial transformations. TAR consists of three main components: a Multi-Scale Visual Feature Learning (MSFL) module, a Text-Assisted Feature Enhancement (TAFE) module, and a Coarse-to-Fine Dense Matching (CFDM) module. 

Firstly, MSFL extracts multi-scale visual features from the optical and SAR images. After that, TAFE first generates remote sensing semantic text features using the frozen RemoteCLIP \cite{liu2024remoteclip}, and then enhances the visual features through visual-text feature interaction and visual-visual feature interaction. Additionally, CFDM establishes coarse correspondences using the enhanced high-level features and refines the matched locations with low-level visual features. Finally, we compute the geometric transformation according to matching correspondences and perform image transformation and registration. The following sections will introduce the details of the network structure and the optimization process.

\subsection{Multi-Scale Visual Feature Learning (MSFL)}
To achieve fast and accurate matching, TAR uses a Multi-Scale Visual Feature Learning (MSFL)  module to extract visual features from optical and SAR images, respectively. Given an input image pair, MSFL employs the ResNet~\cite{he2016deep} backbone with a feature pyramid network to extract multi-scale visual representations. 


For the input optical (${I}_o\in\mathbb{R}^{H\times W}$) and SAR images ${I}_s\in\mathbb{R}^{H\times W}$, their multi-scale visual features can be represented as follows:


\begin{align}
(\mathbf{F}_{o}^{f}, \mathbf{F}_{o}^{c}) = {\mathrm{MSFL}}({I}_o), \\
(\mathbf{F}_{s}^{f}, \mathbf{F}_{s}^{c}) = {\mathrm{MSFL}}({I}_s),
\end{align}
where $\mathrm{MSFL}$ denotes the multi-scale visual feature learning operator. $\mathbf{F}_{o}^{f}\in\mathbb{R}^{H/2\times W/2}$ and $\mathbf{F}_{s}^{f}\in\mathbb{R}^{H/2\times W/2}$ are low-level features of the optical and SAR images, respectively, while $\mathbf{F}_{o}^{c}\in\mathbb{R}^{H/8\times W/8}$ and $\mathbf{F}_{s}^{c}\in\mathbb{R}^{H/8\times W/8}$ are the high-level features with low-resolution. 





\subsection{Text-Assisted Feature Enhancement (TAFE)}
To enhance the visual features of multimodal images, we use the Text-Assisted Feature Enhancement (TAFE) module to insert the remote sensing text semantic priors into visual features. Specifically, TAFE mainly performs visual-text feature enhancement on high-level features for coarse matching. 

Although optical and SAR images present obvious appearance differences for the same land-cover region, they share consistent semantic information related to the image scenes and objects. For example, forests, rivers, farmlands, urban areas, buildings, bridges, and so on. Therefore, remote sensing category text embeddings can serve as modality-agnostic semantic references and reduce ambiguity during correspondence estimation.


We first construct a text prompt set from 224 remote sensing categories. Specifically, we collect a set of basic remote sensing scenes and land-cover objects, such as forest, ocean, farmland, and urban area. To provide richer semantic priors for better support of feature enhancement in diverse remote sensing scenes, GPT is further used to expand these basic remote sensing categories into a broader vocabulary of remote sensing categories. After removing redundant or irrelevant terms, we obtain 224 categories covering natural scenes, artificial structures, and common land-cover objects. For each remote sensing semantic category, a natural-language prompt is generated using the template ``a satellite image of [category]''. Then, TAFE uses the frozen RemoteCLIP text encoder to extract text features from these prompts, which are used for visual feature enhancement. The text features can be represented as follows:
\begin{equation}
\mathbf{F}_{T} = TE(T),
\end{equation}
where $TE$ is the text encoder of the frozen RemoteCLIP, $T$ is the generated text prompts, and $\mathbf{F}_{T}$ is the text features. 

Considering the semantic information in text features is more consistent with the high-level visual feature, TAFE applies the visual and text feature interaction in high-level visual features. Specifically, the text-enhanced visual features are obtained by cross-attention between the text features and visual features:

\begin{equation}
\begin{aligned}
&\mathbf{F}_{o}^{ct}
=
\mathrm{Attn}(\mathbf{F}_{o}^{c},\mathbf{F}_{T}), \\
&\mathbf{F}_{s}^{ct}
=
\mathrm{Attn}(\mathbf{F}_{s}^{c},\mathbf{F}_{T}), \\
\end{aligned}
\end{equation}
where $\mathrm{Attn}$ means attention computation operation, and it can be represented as follows:
\begin{equation}
\begin{aligned}
&\mathrm{Attn}(f_q,f_k)=\mathrm{Softmax}
\left(
\frac{Q{K}^{\top}}{\sqrt{d}}
\right){V},\\
&Q=f_q{W}_q,K=f_k{W}_k,V=f_k{W}_v,
\end{aligned}
\end{equation}
where ${W}_q$, ${W}_k$, and ${W}_v$ are learnable projections, and $d$ denotes the feature dimension.



Additionally, we adopt visual-visual feature interaction to enhance features by modeling both intra-image structural relationships and cross-modal correlations. Specifically, self-attention is first applied to the optical and SAR high-level features to capture contextual dependencies within each image. Then, cross-attention is performed between the optical and SAR features to exchange cross-modal information. This process can be formulated as follows:
\begin{equation}
\begin{aligned}
\bar{\mathbf{F}}_{o}^{c} &= \mathrm{Attn}(\mathbf{F}_{o}^{c}, \mathbf{F}_{o}^{c}),\\
\bar{\mathbf{F}}_{s}^{c} &= \mathrm{Attn}(\mathbf{F}_{s}^{c}, \mathbf{F}_{s}^{c}),\\
\mathbf{F}_{o}^{cv} &= \mathrm{Attn}(\bar{\mathbf{F}}_{o}^{c},\bar{\mathbf{F}}_{s}^{c}),\\
\mathbf{F}_{s}^{cv} &= \mathrm{Attn}(\bar{\mathbf{F}}_{s}^{c},\bar{\mathbf{F}}_{o}^{c}),
\end{aligned}
\end{equation}
where $\bar{\mathbf{F}}_{o}^{c}$ and $\bar{\mathbf{F}}_{s}^{c}$ denote the enhanced visual features of optical and SAR images by single-modal attention learning, respectively, and $\mathbf{F}_{o}^{cv}$ and $\mathbf{F}_{s}^{cv}$ denote the features of optical and SAR images further enhanced by cross-modal attention learning, respectively. 


Finally, we fuse the enhanced high-level visual features from the visual-text feature interaction branch and the visual-visual feature interaction branch for coarse matching. Specifically, the two feature representations are concatenated along the channel dimension and then fed into a three-layer multilayer perceptron (MLP), which consists of three fully connected layers. The first two layers are followed by ReLU activation, while the last layer produces the fused high-level feature. The enhanced high-level visual features can be formulated as follows:
\begin{equation}
\begin{aligned}
\hat{\mathbf{F}}_{o}^{c}
&=
\mathrm{MLP}\left(\mathbf{F}_{o}^{cv}\oplus\mathbf{F}_{o}^{ct}\right), \\
\hat{\mathbf{F}}_{s}^{c}
&=
\mathrm{MLP}\left(\mathbf{F}_{s}^{cv}\oplus\mathbf{F}_{s}^{ct}\right),
\end{aligned}
\end{equation}
where $\oplus$ denotes channel concatenation, $MLP$ is the multilayer perceptron mapping, $\hat{\mathbf{F}}_{o}^{c}$ and $\hat{\mathbf{F}}_{s}^{c}$ are the final enhanced high-level features used for coarse matching. 


The visual-text feature interaction branch provides semantic cues related to remote sensing scenes and land-cover categories, which helps the deep model distinguish ambiguous regions during matching. Additionally, the visual-visual feature interaction branch captures long-range relationships in images and cross-modal correlations between optical and SAR images. By combining these two branches, TAFE can effectively strengthen visual feature representation and improve localization accuracy.

\subsection{Coarse-to-Fine Dense Matching (CFDM)}
After obtaining the enhanced visual features of optical and SAR images, TAR uses the Coarse-to-Fine Dense Matching (CFDM) module to estimate correspondences between the optical and SAR images. CFDM contains two stages: a coarse matching stage based on the high-level features and a fine matching stage based on the low-level features.

\subsubsection{Coarse matching} In the coarse matching stage, CFDM first adds positional encoding to the enhanced high-level features and flattens image features into sequences. Then, CFDM computes the dense similarity between optical and SAR images and performs dual-softmax to obtain the coarse matching confidence matrix:
\begin{equation}
\mathbf{P}(i,j)=\mathrm{Softmax}_i(\mathbf{S}(i,j)) \cdot \mathrm{Softmax}_j(\mathbf{S}(i,j)),
\end{equation}
where $\mathbf{P}(i,j)$ and $\mathbf{S}(i,j)$ denote the coarse matching confidence and the feature similarity between the $i_{th}$ location in the optical image and the $j_{th}$ location in the SAR image, respectively. $\mathrm{Softmax}_i(\cdot)$ and $\mathrm{Softmax}_j(\cdot)$ indicate softmax normalization along the optical and SAR dimensions, respectively. Then, we can obtain coarse matching correspondences ($M_c$) with high confidence, where $\mathbf{P}(i,j)$ exceeds a threshold $\theta_c$. Meanwhile, the potential outliers are further filtered using mutual nearest-neighbor (MNN) checking.



The coarse matching process provides reliable candidate matching locations across images, but their localization accuracy is limited by the low resolution of the high-level features. Therefore, TAR further refines the matching locations in the fine matching stage. 




\subsubsection{Fine matching} 
In the fine matching stage, we first crop local features from the low-level features around coarse matching correspondences with the spatial range of $3\times3$. Let $\mathbf{F}_{o}^{f,w}$ and $\mathbf{F}_{s}^{f,w}$ denote the cropped local low-level features of the optical and SAR images, respectively. These local features are further processed by self-attention (attention learning based on single-modal images) and cross-attention (attention learning between cross-modal images). It can be formulated as follows:
\begin{equation}
\begin{aligned}
\bar{\mathbf{F}}_{o}^{f,w}&= \mathrm{Attn}(\mathbf{F}_{o}^{f,w}, \mathbf{F}_{o}^{f,w}),\\
\bar{\mathbf{F}}_{s}^{f,w}&= \mathrm{Attn}(\mathbf{F}_{s}^{f,w}, \mathbf{F}_{s}^{f,w}),\\
\mathbf{F}_{o}^{fv,w}&= \mathrm{Attn}(\bar{\mathbf{F}}_{o}^{f,w}, \bar{\mathbf{F}}_{s}^{f,w}),\\
\mathbf{F}_{s}^{fv,w}&= \mathrm{Attn}(\bar{\mathbf{F}}_{s}^{f,w}, \bar{\mathbf{F}}_{o}^{f,w}),
\end{aligned}
\end{equation}
where $\bar{\mathbf{F}}_{o}^{f,w}$ and $\bar{\mathbf{F}}_{s}^{f,w}$ denote the local low-level features enhanced by self-attention, and $\mathbf{F}_{o}^{fv,w}$ and $\mathbf{F}_{s}^{fv,w}$ denote the local low-level features further enhanced by cross-attention. 

According to the enhanced local low-level features, we can further refine the matching correspondences by estimating local offsets that maximize the feature similarity between the optical and SAR image patches, thereby obtaining more accurate matching correspondences, $M_f$.

\subsection{Network Optimization}

According to previous work \cite{sun2021loftr}, this paper combines two training losses for optimizing the estimation of initial correspondences in the coarse matching stage and the local refinement of matching positions in the fine matching stage. 

The coarse matching loss aims to constrain the matching confidence of the matching correspondence to be high, while the matching confidence of the non-matching correspondence is low. Considering the imbalance between positive (matching) and negative (non-matching) samples in optical and SAR images, we use focal loss \cite{lin2017focal} in the coarse matching stage. It can be formulated as follows:
\begin{equation}
\begin{aligned}
\mathcal{L}_c
&=
\lambda_{c}^{+}
\sum_{(i,j)\in \mathcal{P}^{+}}
-\alpha_f \left(1-\mathbf{P}(i,j)\right)^{\gamma}
\log \mathbf{P}(i,j) \\
&\quad +
\lambda_{c}^{-}
\sum_{(i,j)\in \mathcal{P}^{-}}
-\alpha_f \mathbf{P}(i,j)^{\gamma}
\log \left(1-\mathbf{P}(i,j)\right),
\end{aligned}
\end{equation}
where $\mathbf{P}(i,j)$ denotes the predicted coarse matching confidence, $\mathcal{P}^{+}$ and $\mathcal{P}^{-}$ denote the sets of positive and negative samples, respectively, $\alpha_f$ and $\gamma$ are hyperparameters, and $\lambda_c^+$ and $\lambda_c^-$ are the weighting coefficients for the positive and negative samples, respectively.  

The fine matching loss aims to minimize the spatial offset error between the predicted matching location and the ground-truth. To improve training stability, we use an uncertainty-weighted $L_2$ loss, which is represented as follows:
\begin{equation}
\mathcal{L}_f =
\frac{1}{|\mathcal{Q}|}
\sum_{k \in \mathcal{Q}}
w_k
\left\|
\mathbf{e}_k^{gt} - \mathbf{e}_k^{pred}
\right\|_2^2,
\end{equation}
where $\mathbf{e}_k^{gt}$ and $\mathbf{e}_k^{pred}$ denote the ground-truth and predicted offsets of the $k_{th}$ sample, respectively, $\mathcal{Q}$ denotes the set of valid fine matching samples, and $w_k$ is the weight of the $k_{th}$ sample, which can be computed from the standard deviation of the feature similarity within the local window in the fine matching stage. This uncertainty-weighted loss can reduce the influence of unreliable matching points during training.



Finally, we combine the coarse matching loss and the fine matching loss for deep model optimization.  The overall loss can be represented as follows:
\begin{equation}
\mathcal{L} = \lambda_c \mathcal{L}_c + \lambda_f \mathcal{L}_f,
\end{equation}
where $\lambda_c$ and $\lambda_f$ denote the weighting coefficients of the coarse matching loss and the fine matching loss, respectively. 
\subsection{Discussion}\label{discussion}
\textit{Why do we adopt text features to enhance visual features?}\\

Optical and SAR images present significant appearance differences due to their distinct imaging mechanisms. When the cross-modal images exhibit a large geometric transformation, the visual features are insufficient for establishing reliable correspondences. The deep model is difficult to deal with image modality changes and geometric deformation, simultaneously. Text semantic priors, which relate to descriptions of remote sensing scenes and land-cover objects, provide modality-invariant information and rotation-invariant information. Text semantic priors can enhance shared visual feature learning from cross-modal images. Additionally, text semantic features have high discriminability for various land-cover objects, which can help the visual model to distinguish non-matching regions and improve the accuracy of matching. Therefore, we utilize text semantic priors to guide visual feature learning, enhancing the invariance and discriminability of visual features, and thereby increasing the stability and robustness of optical and SAR image registration.

\textit{Where can we insert text features into visual features?}\\

In this paper, we merely insert text features into high-level features. Why do we not use the visual-text interaction strategy to enhance low-level features in the fine matching stage? Low-level features mainly present local structures, textural features, and edges, while text features mainly provide high-level semantic information about the remote sensing scene category. The fine matching mainly adopts high-resolution features with rich details to refine the coarse matching estimation and improve matching accuracy. Directly incorporating global semantic information from text features into low-level features may disturb fine localization. Therefore, TAR only performs visual-text interaction to enhance high-level features, which use text guidance information to improve the matching correspondence prediction in the coarse matching stage.

\section{Experiments}\label{experiment}
This section presents the effectiveness of the proposed method for cross-modal remote sensing image registration. We first introduce the dataset, implementation details, and evaluation metrics, and then compare the proposed method with existing state-of-the-art image registration methods in terms of both metric results and visual results. Finally, we conduct ablation studies to analyze the effectiveness of the text-assisted feature enhancement.


\subsection{Datasets}
This paper conducts cross-modal image registration experiments on the SEN1-2 dataset\cite{schmitt2018sen1} and the OSdataset\cite{xiang2020automatic}. 

\begin{figure}[!tbp]
    \centering
    \begin{minipage}{0.11\textwidth}
        \includegraphics[width=\linewidth]{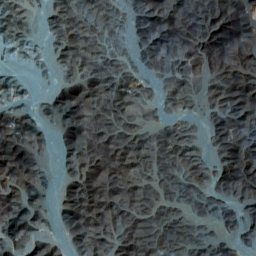}
    \end{minipage}
    \begin{minipage}{0.11\textwidth}
        \includegraphics[width=\linewidth]{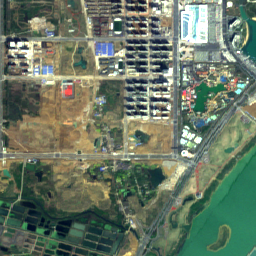}
    \end{minipage}
    \begin{minipage}{0.11\textwidth}
        \includegraphics[width=\linewidth]{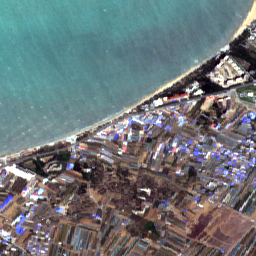}
    \end{minipage}
    \begin{minipage}{0.11\textwidth}
        \includegraphics[width=\linewidth]{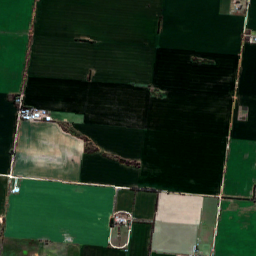}
    \end{minipage}

    \vspace{0.5em}

    \begin{minipage}{0.11\textwidth}
        \includegraphics[width=\linewidth]{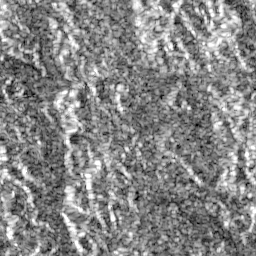}
    \end{minipage}
    \begin{minipage}{0.11\textwidth}
        \includegraphics[width=\linewidth]{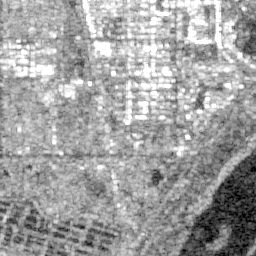}
    \end{minipage}
    \begin{minipage}{0.11\textwidth}
        \includegraphics[width=\linewidth]{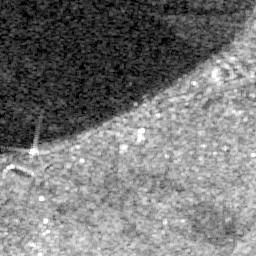}
    \end{minipage}
    \begin{minipage}{0.11\textwidth}
        \includegraphics[width=\linewidth]{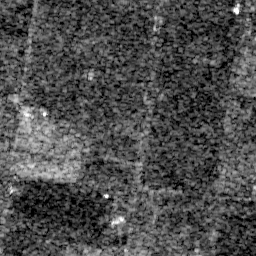}
    \end{minipage}
    \caption{Optical (upper) and SAR (lower) images from the SEN1-2 dataset.}
    \label{fig:datasetsSEN12}
\end{figure}

The SEN1-2 dataset contains a large number of aligned optical and SAR remote sensing image pairs, with an image size of $256 \times 256$ and a spatial resolution of $10\: m$. In total, this dataset consists of 282,384 optical-SAR image patch pairs covering multi-seasons and worldwide, including various remote sensing land-cover objects, such as mountainous areas, urban regions, rivers, and farmlands. In experiments, 12,642 image pairs are randomly selected from the SEN1-2 dataset for training, while another 1,000 image pairs are used for testing. As shown in Fig. \ref{fig:datasetsSEN12}, there are significant appearance differences between optical and SAR images, including grayscale distribution difference, texture structure difference, and the local patterns difference of land-cover objects, which substantially increases the difficulty of cross-modal remote sensing image matching and registration.

\begin{figure}[!tbp]
    \centering
    \begin{minipage}{0.11\textwidth}
        \includegraphics[width=\linewidth]{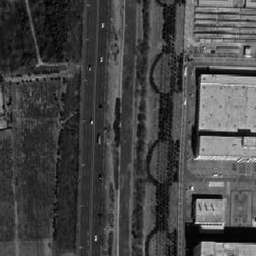}
    \end{minipage}
    \begin{minipage}{0.11\textwidth}
        \includegraphics[width=\linewidth]{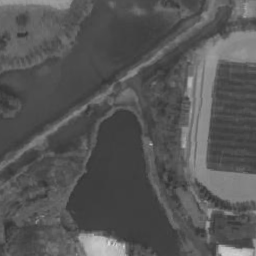}
    \end{minipage}
    \begin{minipage}{0.11\textwidth}
        \includegraphics[width=\linewidth]{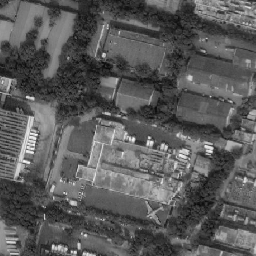}
    \end{minipage}
    \begin{minipage}{0.11\textwidth}
        \includegraphics[width=\linewidth]{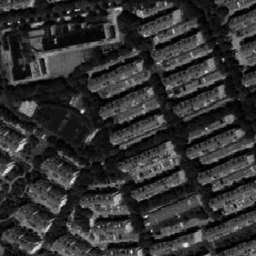}
    \end{minipage}
    
    \vspace{0.5em}
    
    \begin{minipage}{0.11\textwidth}
        \includegraphics[width=\linewidth]{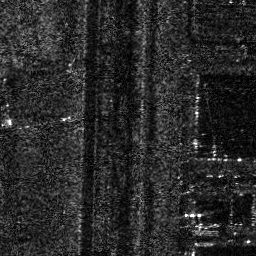}
    \end{minipage}
    \begin{minipage}{0.11\textwidth}
        \includegraphics[width=\linewidth]{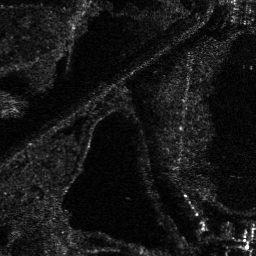}
    \end{minipage}
    \begin{minipage}{0.11\textwidth}
        \includegraphics[width=\linewidth]{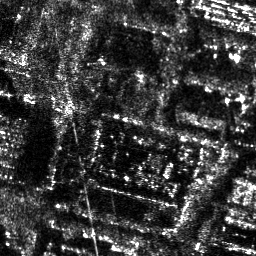}
    \end{minipage}
    \begin{minipage}{0.11\textwidth}
        \includegraphics[width=\linewidth]{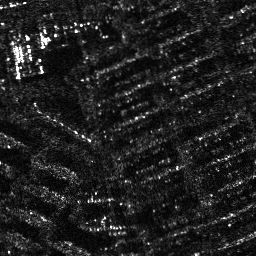}
    \end{minipage}
    \caption{Optical (upper) and SAR (lower) images from the OSdataset.}
    \label{fig:datasetsos}
\end{figure}

The OSdataset contains 10,692 registered optical and SAR image patch pairs, with an image size of $256 \times 256$ and a spatial resolution of $1\: m$. The SAR images are acquired in spotlight mode by the GF-3 satellite, while the optical images are obtained from Google Earth. This dataset covers 20 urban and rural regions worldwide. The image examples of the OSdataset are shown in Fig. \ref{fig:datasetsos}. 
In experiments, 6,297 and 1,000 image pairs are used for training and testing, respectively. 

Compared with SEN1-2, OSdataset has a higher spatial resolution and more complex land-cover details. Therefore, we can validate the effectiveness of the proposed methods on cross-modal remote sensing images with various resolutions. In addition, to simulate the geometric perturbations between images in real-world applications, we apply random geometric transformations to the SAR images. Specifically, affine transformations are used to perturb the SAR images, where the scaling range is set to $[0.7, 1.3]$, the rotation angle range is set to $[-35^\circ, 35^\circ]$, and the translation range is set to 10\% of the image size.

\subsection{Implementation Details}
All experiments are implemented in PyTorch and conducted on an NVIDIA GeForce RTX 3090 GPU. Adam is adopted as the optimizer for model training. The initial learning rate is set to $8\times10^{-3}$, the batch size is set to 8, and the model is trained for 50 epochs. We use a linear warm-up strategy followed by a multi-step learning rate decay, where the learning rate is reduced by a factor of 0.5 at predefined training milestones. In the loss function, the weights of the coarse matching loss and fine matching loss are both set to 1, i.e., $\lambda_c=1$ and $\lambda_f=1$.


\subsection{Evaluation metrics}
In this article, root mean square error (RMSE) and correct match rate ($\mathrm{CMR}$) are adopted as the evaluation metrics for image registration performance. Specifically, RMSE measures the average geometric error between the predicted matched point pairs and the ground-truth correspondences, which reflects the overall localization accuracy of the matching results. It is formulated as follows:
\begin{equation}
\mathrm{RMSE} =
\sqrt{
\frac{1}{N}
\sum_{i=1}^{N}
\left\|
\mathbf{p}_i - \mathbf{p}_i^{gt}
\right\|_2^2
},
\end{equation}
where \(N\) denotes the number of matched point pairs, \(\mathbf{p}_i\) denotes the $i_{th}$ predicted matched point, and \(\mathbf{p}_i^{gt}\) denotes the corresponding ground-truth point.

$\mathrm{CMR}$ measures the successful matching rate under a given error threshold \(\tau\). It is formulated as follows:
\begin{equation}
\mathrm{CMR}@\tau =
\frac{1}{M}
\sum_{j=1}^{M}
\mathbb{I}
\left(
\mathrm{RMSE}_j < \tau
\right).
\end{equation}
where \(\mathbb{I}(\cdot)\) is the indicator function and $M$ denotes the number of test image pairs. This paper presents the $\mathrm{CMR}$ under multiple thresholds, including \(\tau=1\), \(\tau=3\), and \(\tau=5\) pixels, which are denoted as $\mathrm{CMR}@1$, $\mathrm{CMR}@3$, and $\mathrm{CMR}@5$, respectively.






\subsection{Comparison Results}
\begin{table}[!tbp]
\centering
\caption{Comparative results of different methods on the SEN1-2 dataset.}
\label{tab:sen12_compare}
\setlength{\tabcolsep}{8pt}
\renewcommand{\arraystretch}{1.15}
\begin{tabular}{lcccc}
\toprule
\textbf{Methods} & \textbf{RMSE} & \textbf{CMR@1} & \textbf{CMR@3} & \textbf{CMR@5} \\
\midrule
RIFT\cite{li2019rift}    & 17.56 & 0.0  & 0.9  & 9.4  \\
LNIFT\cite{li2022lnift}   & 14.45 & 0.0  & 1.6  & 9.8  \\
OSFlow\cite{xiang2019flow}  & 8.59  & 0.0  & 10.4 & 19.2 \\
ADRNet\cite{xiao2024adrnet}  & 2.33  & 56.1 & 70.7 & 83.8 \\
LoFTR\cite{sun2021loftr}   & 2.43  & \underline{62.8} & 79.4 & 86.0 \\
GDROS\cite{sun2025gdros}   & 1.78  & 47.5 & \underline{83.7} & 88.0 \\
XoFTR\cite{tuzcuouglu2024xoftr}   & \textbf{1.12} & 59.4 & 82.6 & \underline{89.7} \\
Ours    & \underline{1.68} & \textbf{80.3} & \textbf{87.2} & \textbf{90.3} \\
\bottomrule
\end{tabular}
\end{table}

\begin{table}[!tbp]
\centering
\caption{Comparative results of different methods on the OSdataset.}
\label{tab:osdataset_compare}
\setlength{\tabcolsep}{8pt}
\renewcommand{\arraystretch}{1.15}
\begin{tabular}{lcccc}
\toprule
\textbf{Methods} & \textbf{RMSE} & \textbf{CMR@1} & \textbf{CMR@3} & \textbf{CMR@5} \\
\midrule
RIFT\cite{li2019rift}    & 18.54 & 0.0  & 0.3  & 7.3  \\
LNIFT\cite{li2022lnift}   & 14.38 & 0.0  & 1.5  & 9.8  \\
OSFlow\cite{xiang2019flow}  & 10.36 & 0.0  & 9.8  & 16.6 \\
ADRNet\cite{xiao2024adrnet}  & 5.48  & 44.5 & 69.9 & 75.0 \\
LoFTR\cite{sun2021loftr}   & 2.81  & \underline{52.2} & \underline{80.5} & 85.4 \\
GDROS\cite{sun2025gdros}   & 2.16  & 42.6 & 75.1 & 87.0 \\
XoFTR\cite{tuzcuouglu2024xoftr}   & \textbf{1.57} & 32.5 & 79.1 & \underline{87.9} \\
Ours    & \underline{2.03} & \textbf{59.6} & \textbf{84.6} & \textbf{89.2} \\
\bottomrule
\end{tabular}
\end{table}

To validate the effectiveness of the proposed method, we compare our TAR with several representative registration methods, including traditional cross-modal registration methods, i.e., RIFT\cite{li2019rift} and LNIFT\cite{li2022lnift}, as well as deep learning-based methods, i.e., OS-Flow\cite{xiang2019flow}, ADRNet\cite{xiao2024adrnet}, LoFTR\cite{sun2021loftr}, GDROS\cite{sun2025gdros}, and XoFTR\cite{tuzcuouglu2024xoftr} based on the SEN1-2 dataset and OSdataset.

From the overall results in Tables~\ref{tab:sen12_compare} and~\ref{tab:osdataset_compare}, deep learning-based methods significantly outperform traditional registration methods. RIFT, LNIFT, and OS-Flow obtain relatively large RMSE values and low $\mathrm{CMR}$ values on both test datasets. Specifically, they fail to achieve very precise registration accuracy on cross-modal images. In contrast, existing deep learning-based methods, including ADRNet, LoFTR, GDROS, and XoFTR, achieve lower RMSE and higher $\mathrm{CMR}$. This demonstrates the effectiveness and advantages of deep feature learning in cross-modal remote sensing image registration.

Additionally, compared with these deep learning-based methods, our proposed TAR achieves the best $\mathrm{CMR}$ under all thresholds on both datasets. On SEN1-2, TAR obtains $\mathrm{CMR}@1$, $\mathrm{CMR}@3$, and $\mathrm{CMR}@5$ of 80.3\%, 87.2\%, and 90.3\%, respectively. On OSdataset, TAR achieves $\mathrm{CMR}@1$, $\mathrm{CMR}@3$, and $\mathrm{CMR}@5$ of 59.6\%, 84.6\%, and 89.2\%, respectively. In particular, our TAR has significant advantages in precise matching performance over other deep learning methods. For example, the $\mathrm{CMR}@1$ of TAR exceeds  LoFTR, GDROS, and XoFTR by 17.5\%, 32.8\%, and 20.9\% on SEN1-2, respectively. On OSdataset, TAR improves $\mathrm{CMR}@1$ by 7.4\%, 17.0\%, and 27.1\% over LoFTR, GDROS, and XoFTR, respectively. These results show that TAR establishes more reliable correspondences under various localization thresholds. 

It should be noted that although TAR does not achieve the best RMSE, it obtains the best matching performance of CMR under various thresholds. RMSE measures the average localization error and can be affected by a small number of samples with several relatively large residual errors. In contrast, CMR measures the proportion of successfully registered image pairs, which can reflect the stability of matching success under different accuracy requirements. The results indicate that TAR can significantly improve the success rate and stability of optical and SAR registration.

The advantage of TAR mainly comes from the text-assisted feature enhancement in the coarse matching stage. When there are large geometric deformations and obvious appearance differences between optical and SAR images, relying solely on visual features may generate ambiguous coarse correspondences. TAR introduces semantic category priors related to remote sensing scenes and land-cover objects to enhance high-level feature learning, which will help the deep model distinguish potential matching regions during large-range search. Therefore, TAR achieves more stable coarse correspondence estimation and obtains the best CMR across all thresholds.

In summary, the comparison results show that deep learning-based methods are generally more effective than traditional registration methods for optical and SAR image registration. Additionally, TAR further improves the matching stability of deep learning methods by introducing text semantic priors into high-level visual features for coarse matching. These experimental results demonstrate the effectiveness of text-assisted feature learning for optical and SAR image registration.

\subsection{Ablation Study}
\begin{table}[!tbp]
\centering
\caption{Ablation study on the position of text feature interaction on the SEN1-2 dataset.}
\label{tab:ablation_position}
\setlength{\tabcolsep}{6pt}
\renewcommand{\arraystretch}{1.15}
\begin{tabular}{lcccc}
\toprule
\textbf{Methods} & \textbf{RMSE} & \textbf{CMR@1} & \textbf{CMR@3} & \textbf{CMR@5} \\
\midrule
W/O T & \underline{2.11} & \underline{67.4} & \underline{79.2} & \underline{87.5} \\
W/ T, Fine & 2.84           & 49.7           & 61.0            & 72.6            \\
W/ T, Coarse+Fine& 2.45            & 52.6            & 64.3            & 78.7            \\
W/ T, Coarse (Ours) & \textbf{1.68} & \textbf{80.3} & \textbf{87.2} & \textbf{90.3}   \\
\bottomrule
\end{tabular}
\end{table}

\begin{table}[!tbp]
\centering
\caption{Ablation study on the scale of the text semantic descriptors on the SEN1-2 dataset.}
\label{tab:ablation_textscale}
\setlength{\tabcolsep}{7pt}
\renewcommand{\arraystretch}{1.15}
\begin{tabular}{lcccc}
\toprule
\textbf{Methods} & \textbf{RMSE} & \textbf{CMR@1} & \textbf{CMR@3} & \textbf{CMR@5} \\
\midrule
W/O T    & 2.11          & 67.4          & 79.2          & 87.5          \\
W/ T, Basic& \underline{1.86} & \underline{78.1} & \underline{86.4} & \underline{88.7} \\
Ours  & \textbf{1.68} & \textbf{80.3} & \textbf{87.2} & \textbf{90.3} \\
\bottomrule
\end{tabular}
\end{table}


This part verifies the effectiveness of visual-text interaction during the coarse and fine matching stage, and the influence of the scale and description form of the semantic text descriptors. Tables \ref{tab:ablation_position} and \ref{tab:ablation_textscale} report the corresponding experimental results.

\subsubsection{The influence of visual-text interaction}
We first investigate the effect of visual-text interaction at different matching stages, such as without visual-text interaction (W/O T), with visual-text interaction in the fine matching stage  (W/ T, Fine), with visual-text interaction in the coarse matching stage  (W/ T, Coarse), and with visual-text interaction in the coarse and fine matching stages  (W/ T, Coarse+Fine). 

As shown in the table, if we introduce the visual-text interaction into the fine matching stage, the matching results will decrease significantly. When we merely utilize the visual-text interaction in the coarse matching stage, the deep model achieves the best matching performance. Compared with the ``W/O T'', our TAR reduces RMSE by 0.43 and improves $\mathrm{CMR}@1$, $\mathrm{CMR}@3$, and $\mathrm{CMR}@5$ by 12.9\%, 8.0\%, and 2.8\%, respectively. These results demonstrate that text semantic priors are most effective when applied to high-level visual features, which can leverage the global scene information and land-cover semantics to improve coarse matching performance. Meanwhile, the scene category semantic text features are not suitable for low-level visual feature enhancement in the fine matching stage. As we discussed in the Section \ref{discussion}, high-level features and text features contain consistent semantic information, while the low-level features have rich local structures. Thus, we only adopt the semantic text features for high-level visual feature enhancement in the coarse matching stage.

\begin{figure*}[!htbp]
    \centering

    \includegraphics[width=0.24\textwidth]{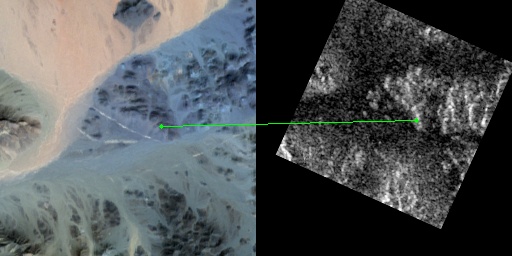}
    \includegraphics[width=0.24\textwidth]{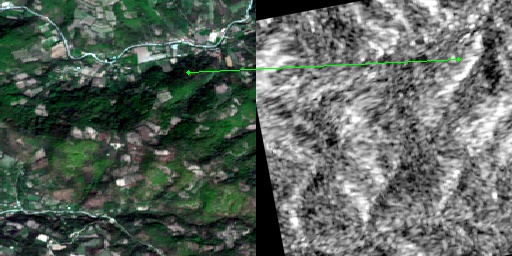}
    \includegraphics[width=0.24\textwidth]{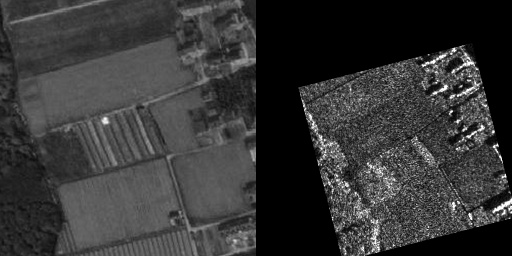}
    \includegraphics[width=0.24\textwidth]{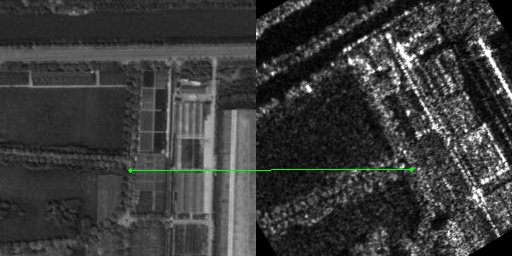}
    \par\vspace{0.2em}
    {\small (a) RIFT}
    \par\vspace{0.6em}

    \includegraphics[width=0.24\textwidth]{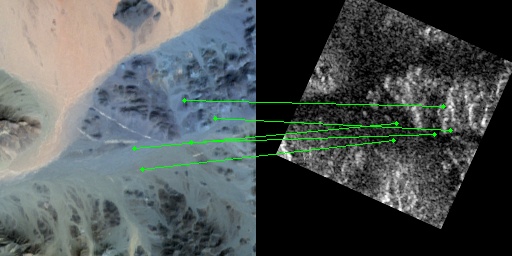}
    \includegraphics[width=0.24\textwidth]{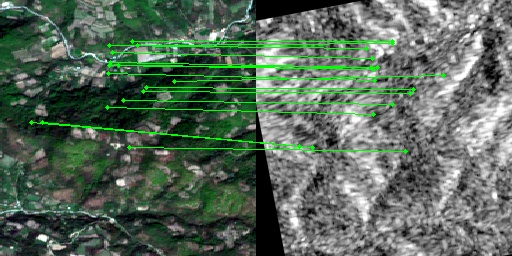}
    \includegraphics[width=0.24\textwidth]{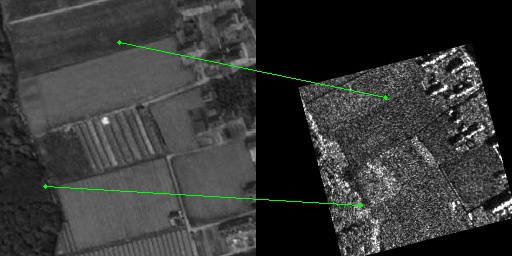}
    \includegraphics[width=0.24\textwidth]{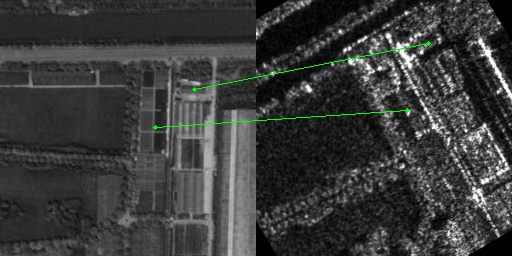}
    \par\vspace{0.2em}
    {\small (b) LNIFT}
    \par\vspace{0.6em}

    \includegraphics[width=0.24\textwidth]{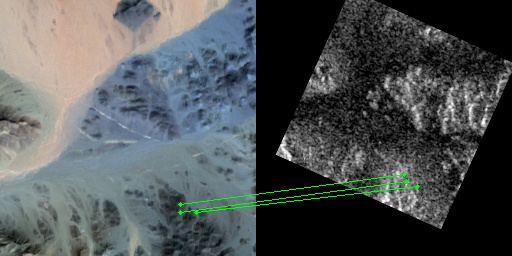}
    \includegraphics[width=0.24\textwidth]{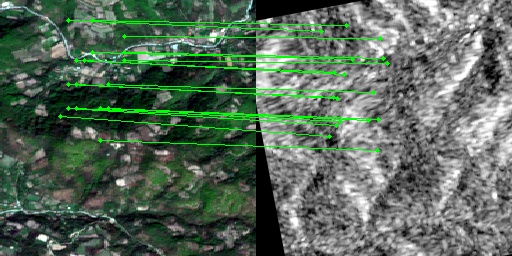}
    \includegraphics[width=0.24\textwidth]{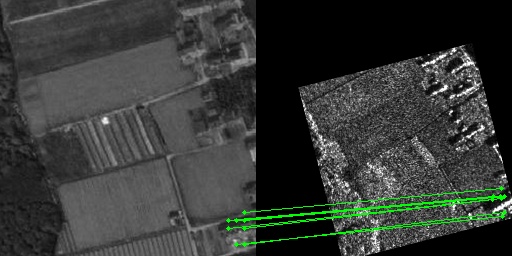}
    \includegraphics[width=0.24\textwidth]{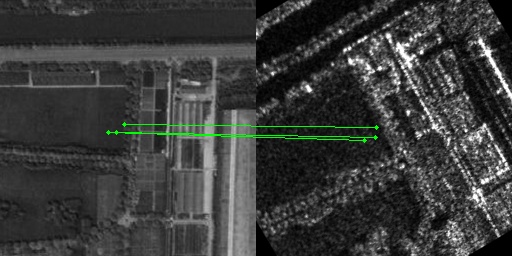}
    \par\vspace{0.2em}
    {\small (c) OS-Flow}
    \par\vspace{0.6em}

    \includegraphics[width=0.24\textwidth]{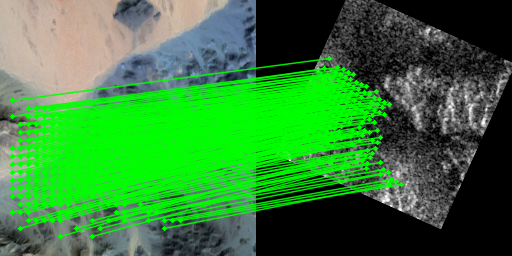}
    \includegraphics[width=0.24\textwidth]{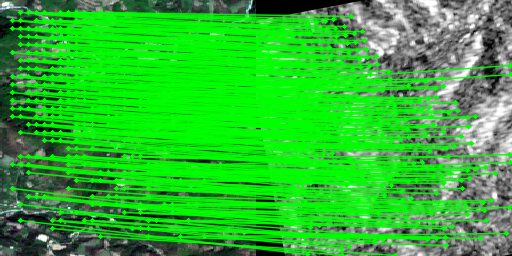}
    \includegraphics[width=0.24\textwidth]{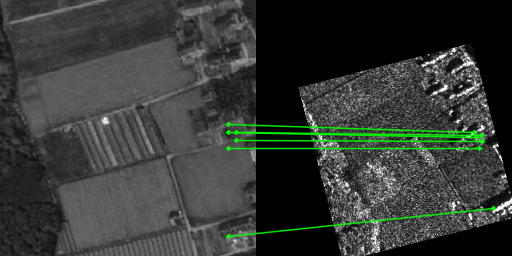}
    \includegraphics[width=0.24\textwidth]{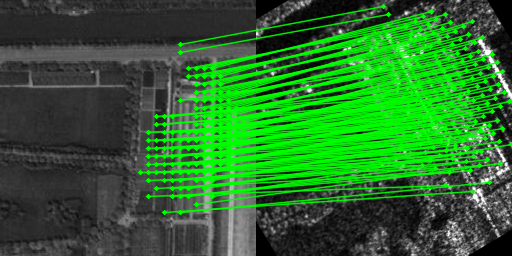}
    \par\vspace{0.2em}
    {\small (d) ADRNet}
    \par\vspace{0.6em}

    \includegraphics[width=0.24\textwidth]{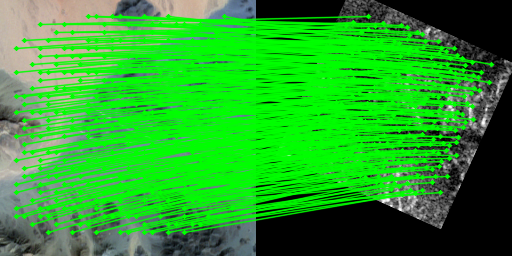}
    \includegraphics[width=0.24\textwidth]{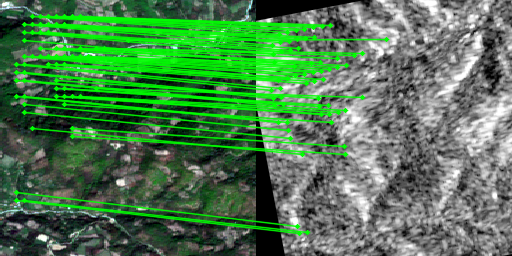}
    \includegraphics[width=0.24\textwidth]{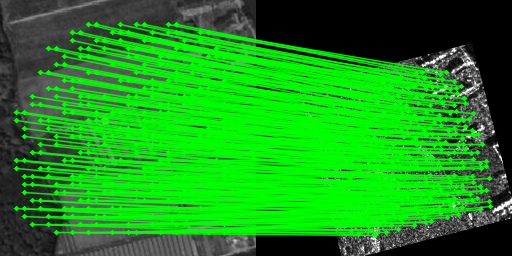}
    \includegraphics[width=0.24\textwidth]{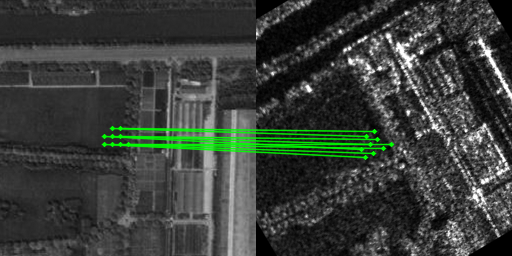}
    \par\vspace{0.2em}
    {\small (e) LoFTR}
    \par\vspace{0.6em}

    \includegraphics[width=0.24\textwidth]{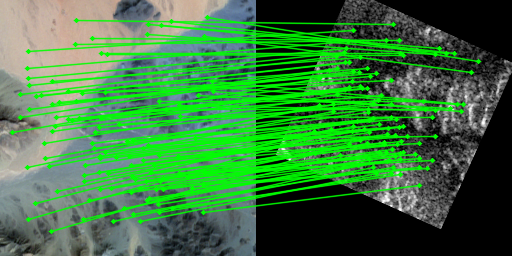}
    \includegraphics[width=0.24\textwidth]{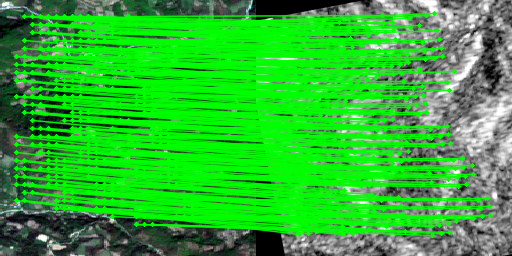}
    \includegraphics[width=0.24\textwidth]{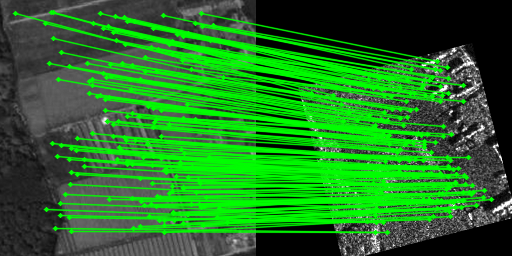}
    \includegraphics[width=0.24\textwidth]{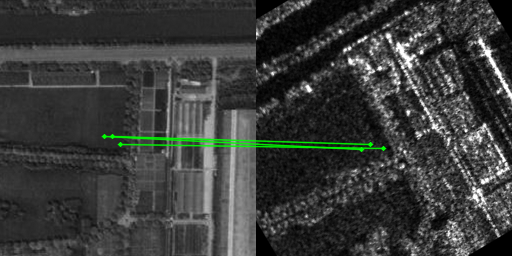}
    \par\vspace{0.2em}
    {\small (f) GDROS}
    \par\vspace{0.6em}

    \includegraphics[width=0.24\textwidth]{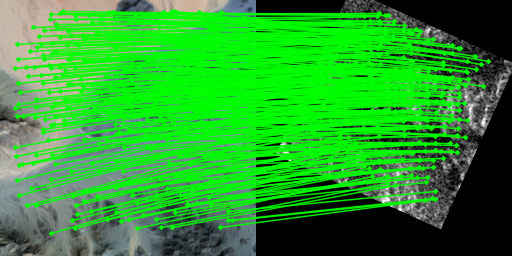}
    \includegraphics[width=0.24\textwidth]{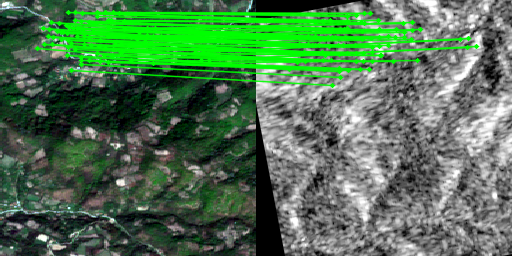}
    \includegraphics[width=0.24\textwidth]{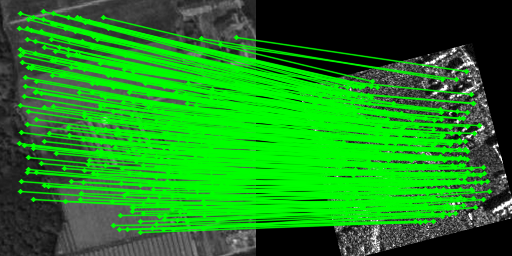}
    \includegraphics[width=0.24\textwidth]{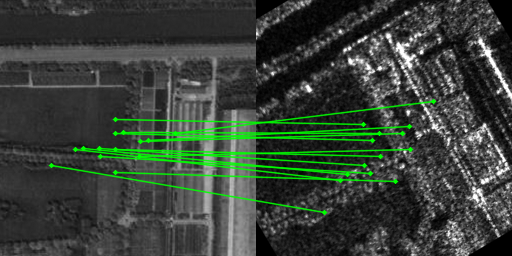}
    \par\vspace{0.2em}
    {\small (g) XoFTR}
    \par\vspace{0.6em}

    \includegraphics[width=0.24\textwidth]{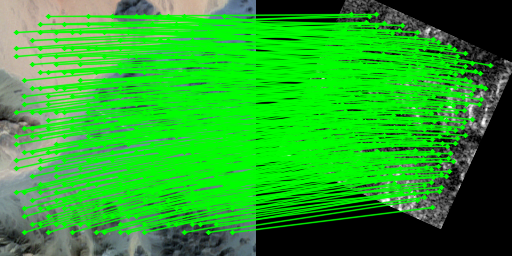}
    \includegraphics[width=0.24\textwidth]{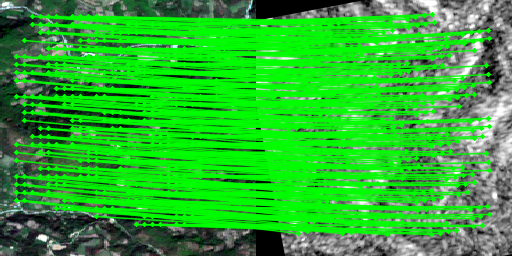}
    \includegraphics[width=0.24\textwidth]{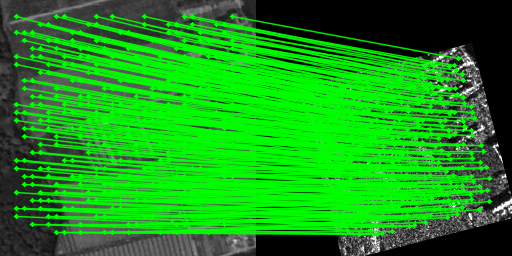}
    \includegraphics[width=0.24\textwidth]{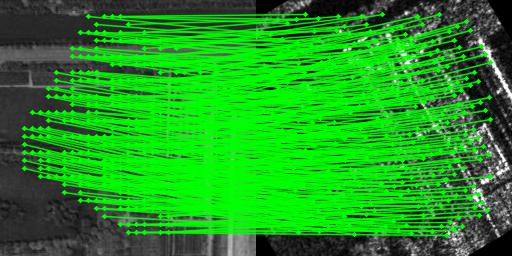}
    \par\vspace{0.2em}
    {\small (h) TAR (Ours)}

    \caption{The visualized matching results of different methods on optical and SAR images. The green lines indicate successful matching point pairs, which have a spatial offset error that is smaller than 3 pixels.}
    \label{fig:qualitative_results}
\end{figure*}


\subsubsection{The influence of the text descriptions}
This part tests the influence of the scale of the text descriptions. In Table~\ref{tab:ablation_textscale}, we compare the matching results of the deep model without semantic text information for visual feature enhancement (W/O T), the deep model with only 37 basic semantic categories for building text descriptions (W/ T, Basic), and our proposed TAR, which contains 224 remote sensing categories for text-assisted visual feature learning.

Compared with the ``W/O T'', the deep model with semantic text descriptions achieves significant matching performance improvements. For example, the RMSE of ``W/O T'' decreases from 2.11 to 1.86 by ``W/ T, Basic'', while $\mathrm{CMR}@1$, $\mathrm{CMR}@3$, and $\mathrm{CMR}@5$ increase from 67.4\%, 79.2\%, and 87.5\% to 78.1\%, 86.4\%, and 88.7\%, respectively. These results indicate that even a compact set of remote sensing semantic categories can provide useful high-level semantic priors for boosting optical and SAR matching. When we use the rich semantic text descriptors with 224 categories, the matching results can be further improved. Thus, the broader remote sensing categories provide richer semantic information, enabling the deep model to identify the matching location during the coarse matching stage.

\subsection{Visualization Results}
To provide a more intuitive comparison of different methods on optical and SAR image registration, we present the visual matching results in Fig.~\ref{fig:qualitative_results}. The green lines denote successful matching point pairs with a spatial offset error that is smaller than 3 pixels.

As shown in Fig.~\ref{fig:qualitative_results}, RIFT, LNIFT, and OS-Flow acquire relatively sparse and unstable matching results. RIFT and LNIFT rely on handcrafted local descriptors, which cannot extract discriminative features when there are obvious appearance differences and large geometric transformations across images. Deep learning-based methods, including ADRNet, LoFTR, GDROS, and XoFTR, generally produce more reliable matching correspondences than traditional methods. However, these deep learning methods still fail to acquire sufficient local correspondences in some challenging samples. Our proposed TAR can obtain dense and uniformly distributed matching correspondences on various optical and SAR images. These visual matching results also demonstrated the effectiveness and advantages of TAR, which leverages semantic text priors to enhance visual features, thereby establishing more reliable correspondences and obtaining more stable matching results in cross-modal image registration.

\section{Conclusion}\label{conclusion}
This paper proposes TAR, a text semantic-assisted framework for optical and SAR image registration. TAR mainly utilizes the text semantic prior information to enhance high-level visual feature learning and obtain more accurate registration. It contains three modules: a multi-scale visual feature learning (MSFL) module to extract low-level and high-level visual features from remote sensing images, a text-assisted feature enhancement (TAFE) module to enhance high-level visual feature learning by visual-text interaction, and a coarse-to-fine dense matching (CFDM) module to identify the matching correspondences based on the multi-scale visual features. Experiments on the SEN1-2 and OSdataset datasets demonstrate the effectiveness and advantages of our TAR. TAR achieves the best correct match rates under multiple error thresholds and maintains competitive RMSE values. The text semantic priors improve the reliability of coarse correspondence estimation and increase the matching accuracy of optical and SAR images. In future work, we will explore more effective text semantic prompts and visual-text interaction methods for remote sensing image registration.

\ifCLASSOPTIONcaptionsoff
\newpage
\fi
\bibliographystyle{IEEEtran}
\bibliography{references}
\begin{IEEEbiography}[{\includegraphics[width=1in,height=1.25in,clip,keepaspectratio]{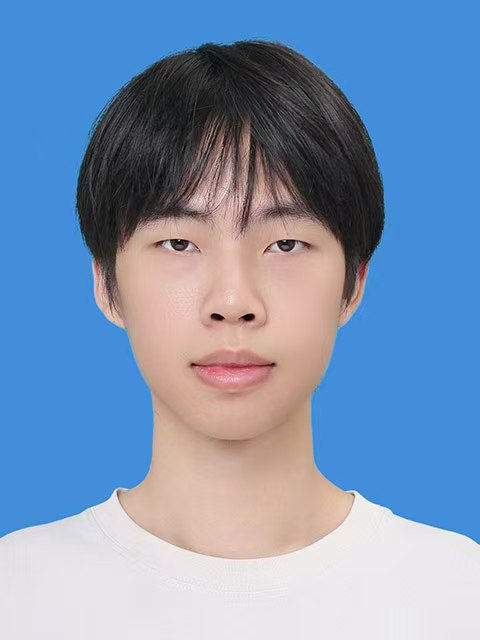}}]{Zhuoyu Cai} received the B.S. degree in Artificial Intelligence from Xidian University, Xi'an, China, in 2025. He is currently pursuing the M.S. degree with the Key Laboratory of Intelligent Perception and Image Understanding of Ministry of Education, School of Artificial Intelligence, Xidian University, Xi'an, China.

His research interests include remote sensing image registration.
\end{IEEEbiography}

\begin{IEEEbiography}[{\includegraphics[width=1in,height=1.25in,clip,keepaspectratio]{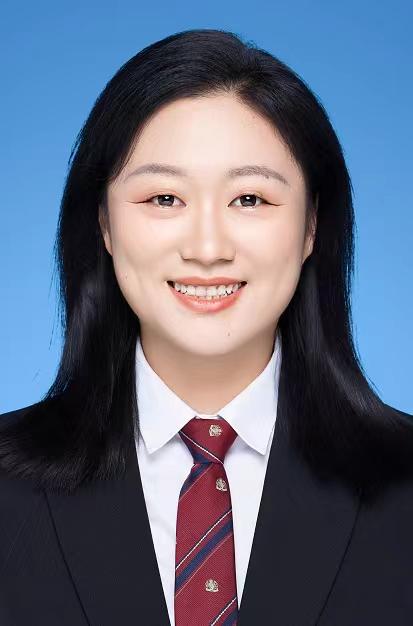}}]{Dou Quan} received the B.S. degree in 2015 and the Ph.D. degree in 2021, from Xidian University, Xi’an, China. She is currently an associate professor with the Key Laboratory of Intelligent Perception and Image Understanding of Ministry of Education of China, School of Artificial Intelligence, Xidian University.
		
Her research interests include machine learning, deep learning, metric learning, image matching, image registration, and image classification.
\end{IEEEbiography}

\begin{IEEEbiography} [{\includegraphics[width=1in,height=1.25in,keepaspectratio]{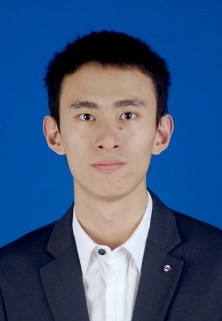}}]{Ning Huyan} received his B.S. degree in 2015 and Ph.D. degree in 2022 from Xidian University, Xi’an, China. From 2019 to 2020, he was a joint Ph.D. candidate under the supervision of Prof. Jocelyn Chanussot at the Inria Grenoble Rhône-Alpes research center in France. From 2022 to 2024, he worked as a researcher at SenseTime in Xi’an, China. Currently, he is a postdoctoral researcher in the Department of Automation at Tsinghua University. 
		
His research interests include hyperspectral image anomaly detection, human pose estimation, and motion generation.
\end{IEEEbiography}
	
\begin{IEEEbiography}[{\includegraphics[width=1in,height=1.25in,clip,keepaspectratio]{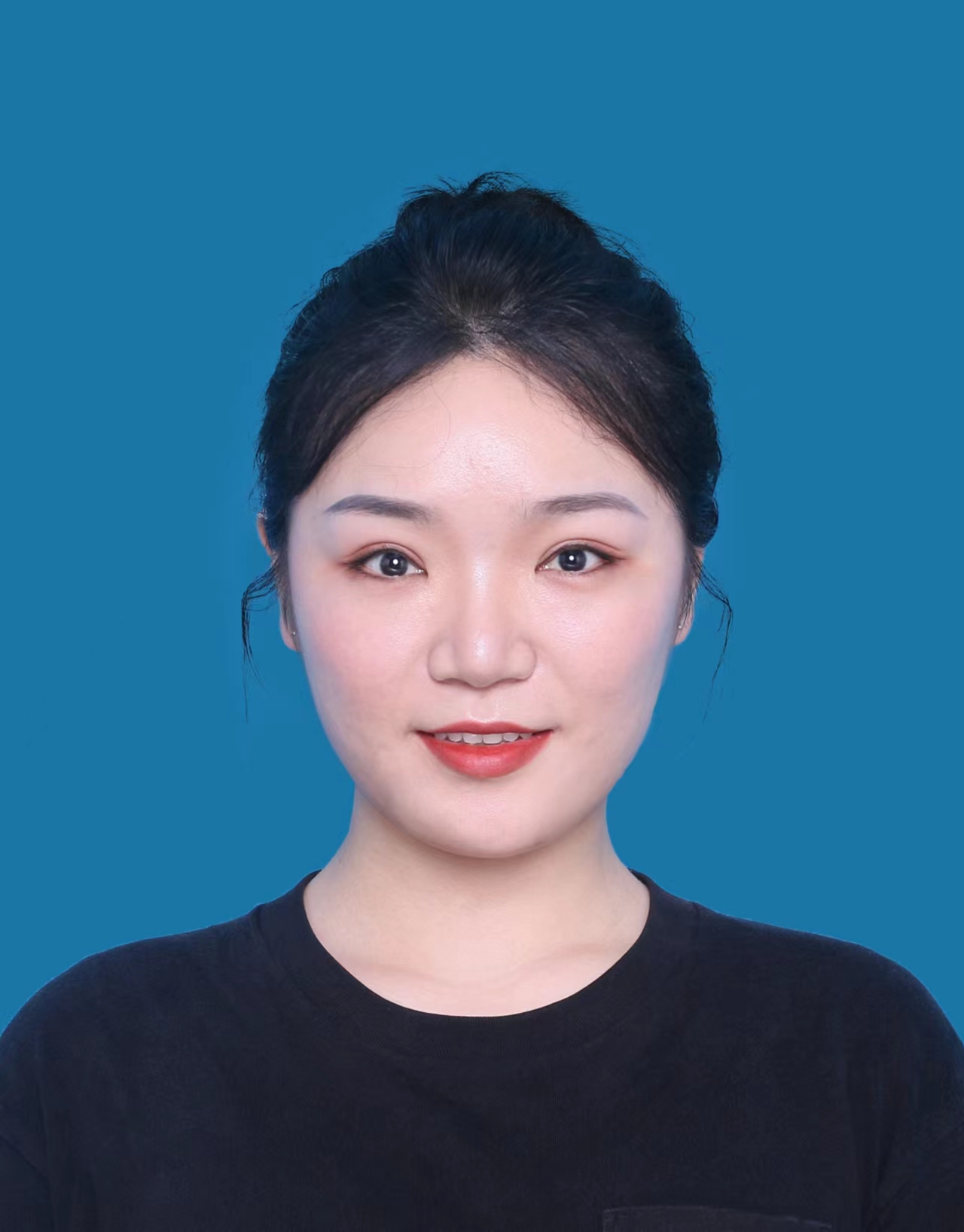}}]{Pei He} received the B.S. degree in information engineering and the Ph.D. degree in computer science and technology from Xidian University, Xian, China, in 2018 and 2025, respectively. She is currently a postdoctoral researcher in the School of Artificial Intelligence at Xidian University. 
		
Her current research interests include computer vision, semantic segmentation, and generalized representation.
\end{IEEEbiography}
	
\begin{IEEEbiography} [{\includegraphics[width=1in,height=1.25in,keepaspectratio]{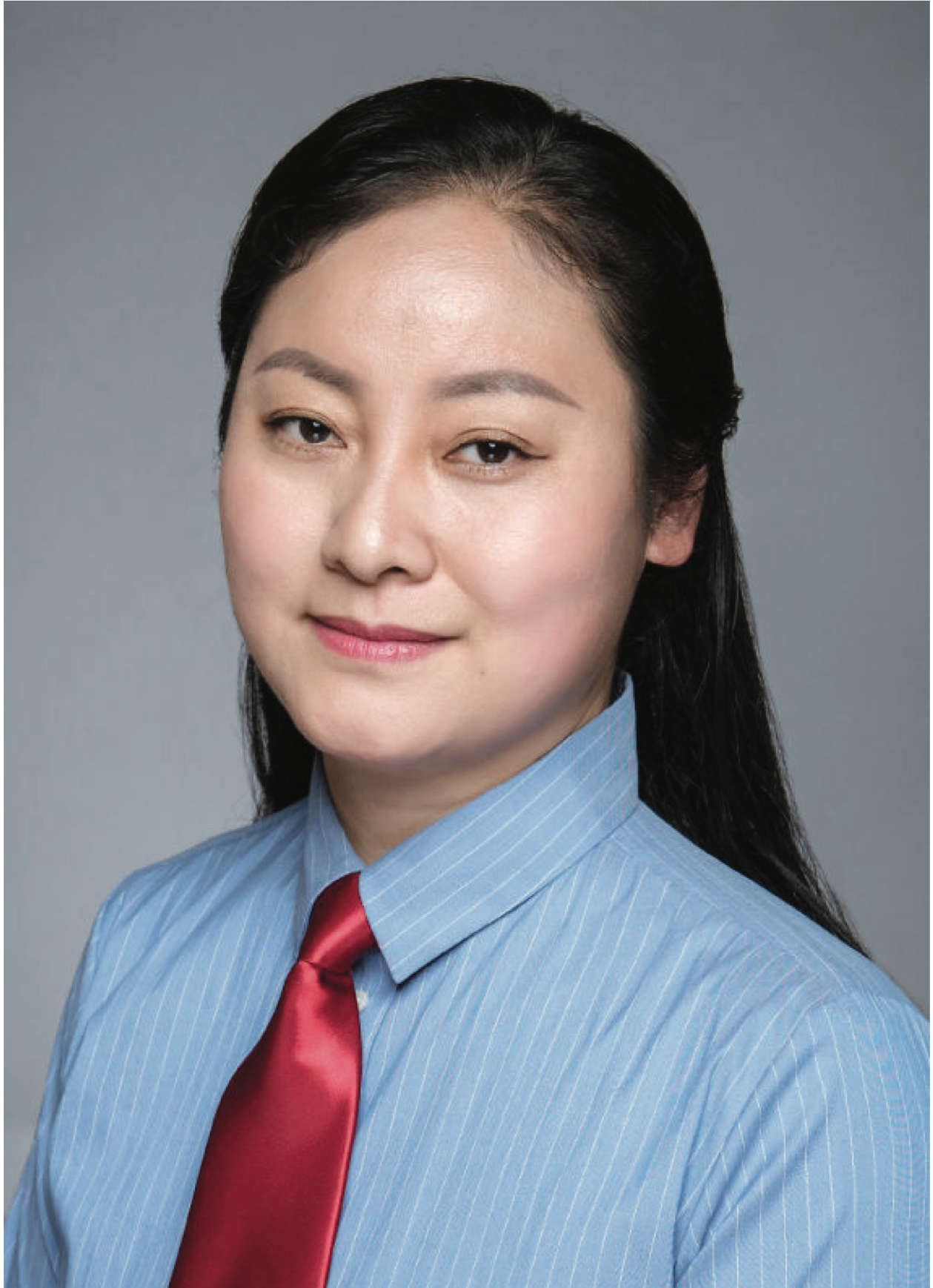}}]{Shuang Wang} received the B.S. degree in 2000, the M.S. degree in 2003, and the Ph.D. degree in circuits and systems in 2007 from Xidian University, Xi'an, China. She is currently a Professor with the Key Laboratory of Intelligent Perception and Image Understanding of Ministry of Education of China, Xidian University.
		
Her research interests include sparse representation, image processing, synthetic aperture radar (SAR) automatic target recognition, remote sensing image captioning, and polarimetric SAR data analysis and interpretation.
\end{IEEEbiography}

\begin{IEEEbiography}[{\includegraphics[width=1in,height=1.25in,clip,keepaspectratio]{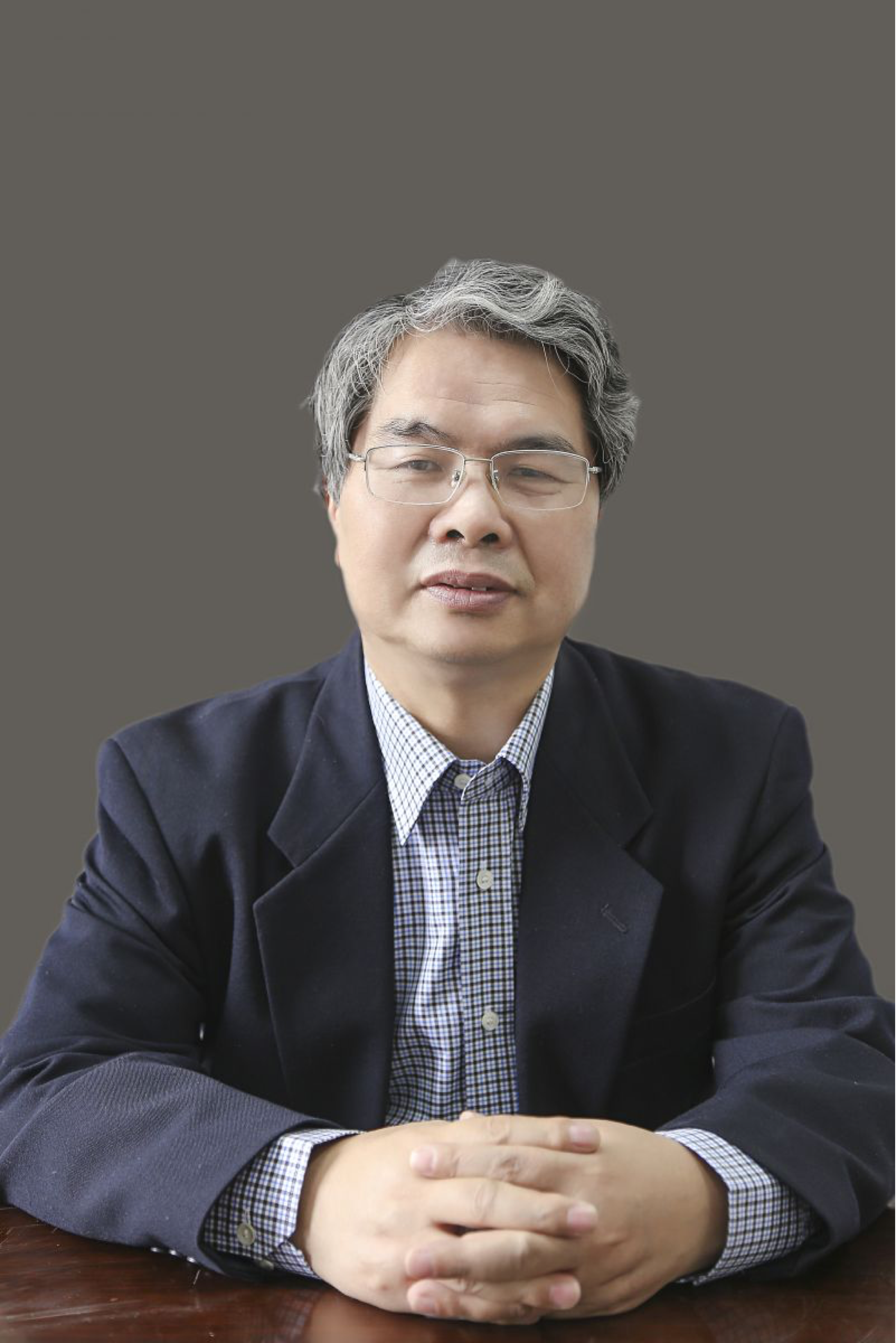}}]{Licheng Jiao} was born in Shaanxi, China, on October 15, 1959. He received the B.S. degree from Shanghai Jiaotong University, China, in 1982 and the M.S. and Ph.D. degrees from Xi'an Jiaotong University, Xi'an, China, in 1984 and 1990, respectively. From 1984 to 1986, he was an Assistant Professor with the Civil Aviation Institute of China, Tianjin, China. During 1990 and 1991, he was a Postdoctoral Fellow with the Key Lab for Radar Signal Processing, Xidian University, Xi'an, China. Currently, he is the Director of the Key Laboratory of Intelligent Perception and Image Understanding of the Ministry of Education of China.
		
His current research interests include signal and image processing, nonlinear circuits and systems theory, learning theory and algorithms, optimization problems, wavelet theory, and machine learning.
	\end{IEEEbiography}
\end{document}